# Clarifying Trust of Materials Property Predictions using Neural Networks with Distribution-Specific Uncertainty Quantification

Cameron J. Gruich[a,b], Varun Madhavan[a], Yixin Wang[c], and Bryan R. Goldsmith[a,b,*]

[a] Department of Chemical Engineering, University of Michigan, Ann Arbor, Michigan 48109-2136, USA

[b] Catalysis Science and Technology Institute, University of Michigan, Ann Arbor, Michigan 48109-2136, USA

[c] Department of Statistics, University of Michigan, 1085 S University Ave, Ann Arbor, 48109-1107, MI, USA

**ABSTRACT:** It is critical that machine learning (ML) model predictions be trustworthy for high-throughput catalyst discovery approaches. Uncertainty quantification (UQ) methods allow estimation of the trustworthiness of an ML model, but these methods have not been well explored in the field of heterogeneous catalysis. Herein, we investigate different UQ methods applied to a crystal graph convolutional neural network (CGCNN) to predict adsorption energies of molecules on alloys from the Open Catalyst 2020 (OC20) dataset, the largest existing heterogeneous catalyst dataset. We apply three UQ methods to the adsorption energy predictions, namely $k$-fold ensembling, Monte Carlo dropout, and evidential regression. The effectiveness of each UQ method is assessed based on accuracy, sharpness, dispersion, calibration, and tightness. Evidential regression is demonstrated to be a powerful approach for rapidly obtaining tunable, competitively trustworthy UQ estimates for heterogeneous catalysis applications when using neural networks. Recalibration of model uncertainties is shown to be essential in practical screening applications of catalysts using uncertainties.

## 1. INTRODUCTION

Machine learning (ML) approaches have rapidly grown in popularity to accelerate catalyst screening and understanding [1–4]. Making predictions of catalyst properties with ML models is orders of magnitude faster compared to first-principles simulation of catalysts, for example, using density functional theory (DFT) modeling to compute adsorption energies of molecules. DFT modeling combined with ML has emerged as a compelling approach for rapid materials characterization, enabling a several-orders-of-magnitude expansion in the number of materials able to be studied compared to DFT modeling alone [5–7]. However, the potential of ML models to efficiently explore large catalyst spaces can only be achieved if it is simple to detect when ML model predictions are accurate or highly uncertain.

Uncertainty provides a means to infer the accuracy of a model without explicitly knowing the accuracy, but often this uncertainty of the ML model cannot be trusted [8]. Reliable uncertainty quantification (UQ) of ML model predictions is a fundamental challenge to ML-guided materials and molecule discovery [8–12]. Predictive uncertainty can be estimated with either distribution-specific methods or distribution-free methods (e.g., Monte Carlo dropout [13] with a Gaussian-specific calibration assumption [14] or conformal UQ [15] respectively). By distribution-specific methods, we mean methods that assume an underlying distribution shape to the uncertainty of model predictions or their calibrated scaling, while distribution-free methods make no such assumption. Established UQ methods for ML models are often costly to obtain and have limitations in assessing prediction errors for chemical space exploration [16].

Effective uncertainty estimates of predictions are an important design element to enhance several advanced ML strategies for materials discovery, such as high-throughput screening [17,18], transfer learning [19], and active learning [20,21], as illustrated in **Fig. 1**. For catalyst discovery, one can ascertain if an ML model is making accurate predictions of catalyst behavior by confirming the predicted properties experimentally or through first-principles calculations. In a high-throughput context where many thousands of systems or more are being studied, frequently confirming the predictive accuracy of the ML model via experiment or first-principles calculations is impractical. UQ is a means by which one can quantify the trustworthiness of ML predictions in a way that is practical and attainable for materials discovery strategies that explore vast materials spaces.

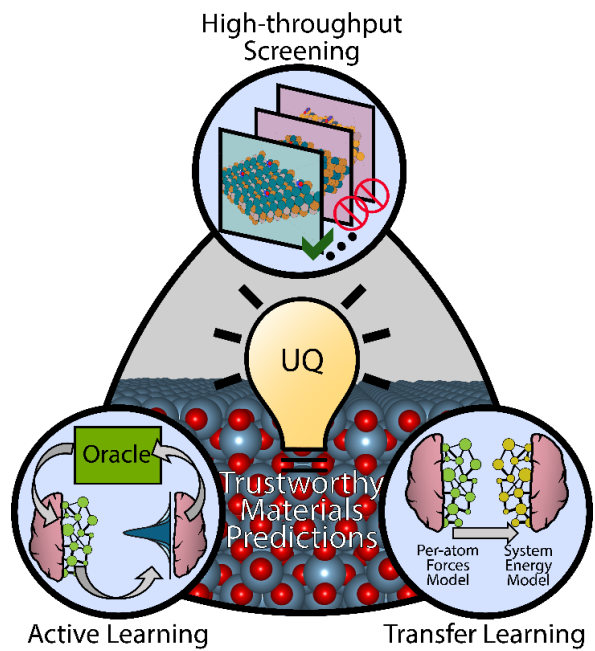

**Fig. 1. Advanced materials discovery strategies are enabled with uncertainty quantification**. Screening, active learning, and transfer learning are enhanced by trustworthy estimates of predictive uncertainty, particularly in high-throughput applications where the size of the uncertainty estimate is used to infer model accuracy. The oracle refers to some trustworthy system that outputs the desired target, such as DFT-accurate materials properties.

ML-guided studies of catalysts that apply uncertainty have already been performed, but these studies historically rely on Gaussian process regression [22–25]. Gaussian process regression is formulated from Bayesian statistics and directly outputs uncertainty estimates of the predictions [26]. However, the computational cost of training a Gaussian process regression model typically scales $O(N^3)$ and thus grows unfavorably with dataset size, which is an open challenge for big-data applications [27]. Therefore, there is a practical limitation of these models for high-throughput discovery strategies.

Neural networks (NNs) are popular because of their relatively high accuracy across diverse chemical spaces and excellent performance for large datasets [28–32]. For the Open Catalyst 2020 (OC20) dataset [33], state-of-the-art deep NN model variants (e.g., CGCNN [34], SchNet [35], and DimeNet++ [36]) have shown up to a 40% improvement in their accuracy of adsorption energy predictions by using 460,000 training samples versus 10,000 samples [33]. The arrival of large training datasets—and innovations in NN model architectures—has led to steady improvements in accuracy for catalyst property predictions, such as molecular adsorption energies [37]. However, less attention in the field has been devoted to understanding and quantifying the predictive uncertainty associated with these NN architectures applied to prevailing catalysis datasets [14,15].

For uncertainty-guided sample acquisition and efficient model training for catalysis using NNs, it is crucial to ensure that the uncertainty estimates from a given UQ method are reliable and useful. Typical metrics for evaluating uncertainty estimates include accuracy, sharpness, dispersion, calibration, and tightness, which are discussed in Section 2.4. The reliability of a UQ estimate is often primarily discussed in terms of calibration, which refers to how well the uncertainty estimate represents a confidence interval that encloses the ground-truth target [38–43]. Calibrated uncertainty estimates can be used to forecast the accuracy of materials predictions, thereby quantifying the trustworthiness of an ML model to examine and propose materials for further study. Using calibrated uncertainty estimates to forecast whether a large set of ML predictions are reliably accurate or not is common to many applications, such as autonomous driving [44], image analysis [45], and materials discovery [46].

Herein, we compare three different UQ methods, namely, $k$-fold ensembling [47–49], Monte Carlo (MC) dropout [13], and evidential regression [50,51] to better ensure effective use of NN model architectures for high-throughput catalyst discovery strategies. All three methods are able to express uncertainty as a standard deviation ($\sigma$). Evidential regression has seen use in small molecule prediction [50] but has not yet been explored for catalysis applications to our knowledge. These three UQ methods are studied using a crystal graph convolutional neural network (CGCNN) to predict adsorption energies of molecules on solid catalyst materials based on the OC20 dataset, many of which are binary and ternary alloys [33]. Alloy catalysts are highly relevant in industrial applications, such as the Haber-Bosch process [52], catalytic cracking of hydrocarbons [53], and naphtha reforming [54]. The OC20 is an ideal test case for understanding UQ methods because it is the largest and most diverse heterogeneous catalysis dataset, which represents a case study for many material search challenges.

We use accuracy, sharpness, dispersion, calibration, and tightness as UQ metrics to compare the trustworthiness of each of the three UQ methods, **Fig. 2**. We define a trustworthy UQ method as one that is perfectly calibrated because calibration measures how well the uncertainty intervals probabilistically grow or shrink with the predictive error. In practice, one may not know if a UQ method is perfectly calibrated; moreover, it is possible for two UQ methods to have the same average measure of calibration but have different uncertainty estimates for the same materials, so accuracy, sharpness, dispersion, and tightness are discussed as secondary trustworthiness criteria when comparing UQ methods. We particularly emphasize differences in calibration between UQ methods via adversarial group calibration [55] to determine their appropriateness for high-throughput catalyst discovery strategies. Scalar recalibration [56] is applied to address poor calibration performance between UQ methods. Evidential regression is found to be competitively trustworthy before recalibration and the most trustworthy after recalibration. We demonstrate the use of UQ and evidential regression to enumerate materials for DFT-predicted adsorption in the case of a hydrogen adsorbate. Improved reliability in the uncertainty estimates for this demonstration is observed after recalibration. This work will guide future efforts of uncertainty-guided catalyst search and discovery using ML.

## 2. EXPERIMENTAL METHODS

### 2.1 Dataset: Open Catalyst 2020

We used the OC20 dataset to train and test our CGCNN model to predict adsorption energies of molecules on catalyst surfaces [33]. OC20 encompasses a state-of-the-art collection of DFT calculations of adsorption energies on binary and ternary alloy catalysts spanning the periodic table, as well as pure metals. The dataset is comprised of 82 nitrogen, oxygen, and carbon-containing adsorbates. OC20 has over 130,000,000 data points associated with 5,243 unique alloy catalyst compositions.

Within OC20, there are different versions of datasets depending on what task an ML model is performing to predict adsorption energy. We performed the initial-structure-to-relaxed-energy (IS2RE) task, where the ML model accepts an unrelaxed initial structure of an adsorbate/alloy system and predicts the relaxed energy of the system. Electronic energies of the geometry-optimized (i.e., relaxed) gas phase species and the bare alloy surface are subtracted from the combined adsorbate/alloy system electronic energy to obtain an adsorption energy at 0 K [33]. We trained the CGCNN models on the entire IS2RE training dataset composed of 460,328 systems. To benchmark the model accuracy and the effectiveness of the UQ methods, test adsorption energy predictions were made on the entire IS2RE Validation In-Domain (Val-ID) dataset composed of 24,943 systems. The Val-ID dataset was composed of adsorbate/alloy systems that the model has never seen before during training. We herein refer to the Val-ID dataset as the test dataset, and we use this dataset as the test dataset because the dedicated test dataset provided within OC20 does not include target labels due to its usage in leaderboard competitions.

### 2.2 Model: CGCNN

We used a CGCNN model for adsorption energy predictions [33,34,57]. The CGCNN in our study is a deep-learning convolutional NN of atomic surface structures using the distances between atoms in the crystal structure as encoded edge information. Gaussian basis functions are used to encode the distances between atoms [33]. Descriptions of unique bonds between atoms (i.e., atom embeddings) and the distances between atoms (i.e., distance embeddings) were computed. The atom and distance embeddings were combined with element-specific properties of atoms in the system [33,34]. These combined input features pass through convolutional layers that reduce the incoming high-dimensional vector to a fixed lower-dimensional vector. The model accepts adsorbate/alloy systems of a varying number of atoms because all the data in transit inside the model gets convolved to the same number of dimensions. After the convolutional layers, the data was transformed by a series of fully connected layers. Model parameters were optimized to produce accurate predictions via mini-batch stochastic gradient descent. The best reported hyperparameters reported in the original OC20 dataset report were used for our CGCNN model [33]. Before applying any UQ method, our baseline CGCNN model had a 0.651 eV mean absolute error (MAE) on the IS2RE Val-ID dataset, which is comparable to the expected error reported in the literature [33].

### 2.3 Uncertainty Quantification Methods

We compared three different UQ methods from the literature: $k$-fold ensembling [47–49], MC dropout

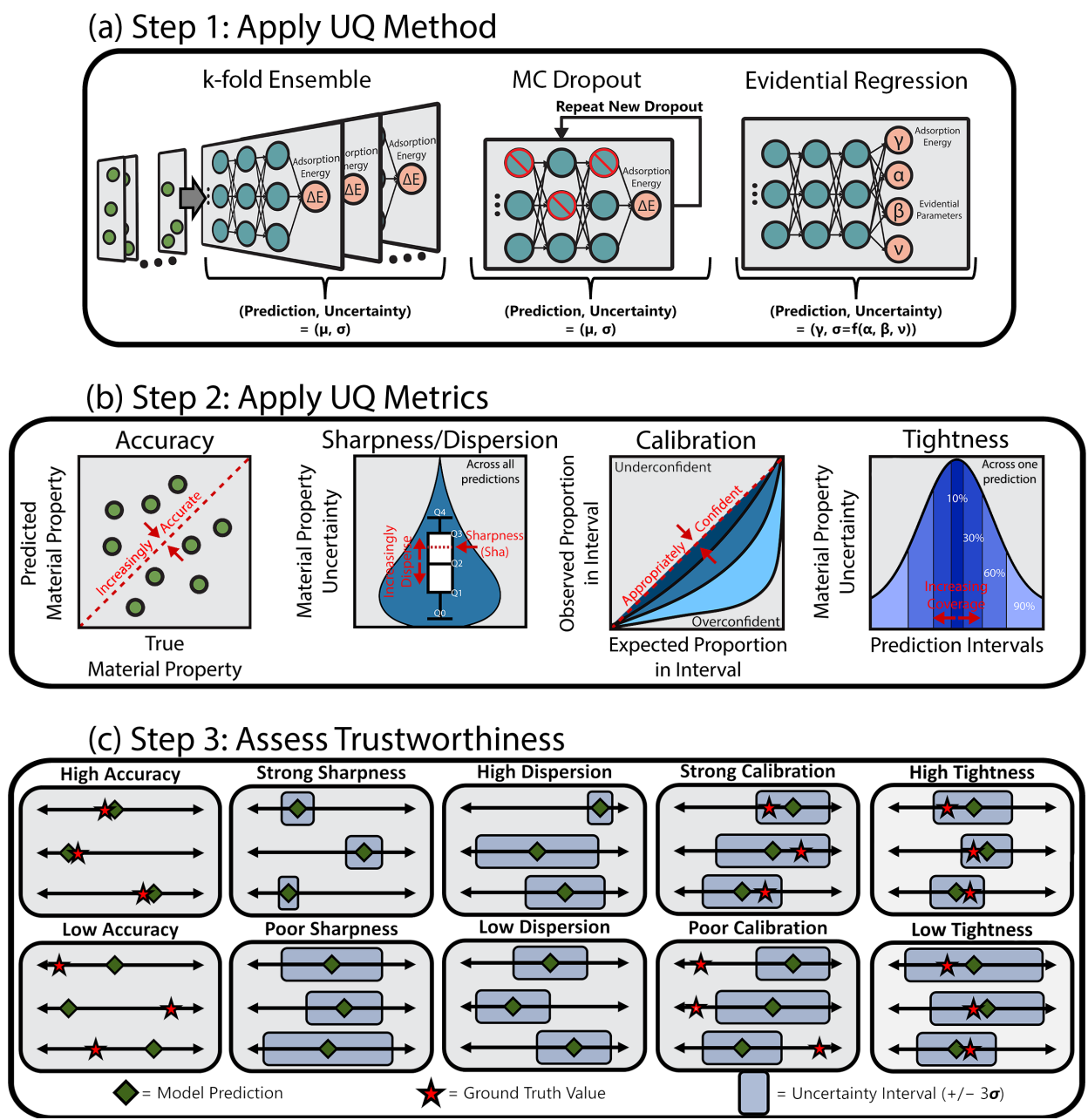

**Fig. 2. Uncertainty quantification workflow to predict adsorption energies of molecules on catalysts. (a)** Three UQ methods (left) *k*-fold ensemble method, (middle) Monte Carlo dropout, and (right) evidential regression are applied onto a crystal graph neural network to give both a property prediction and associated uncertainty. **(b)** UQ metrics of accuracy, sharpness, dispersion, calibration, and tightness are applied to compare UQ methods. **(c)** UQ metrics are interpreted to assess the trustworthiness of materials predictions.

[13], and evidential regression [50,51], **Fig. 2a**. Each adsorption energy prediction was expressed as a mean prediction $\mu$ with uncertainty as a standard deviation $\sigma$ centered around the prediction. Uncertainty intervals around each prediction were constructed using the associated value of the standard deviation $\sigma$ (i.e., an interval of $\mu \pm 3\sigma$). From an application standpoint, these uncertainty intervals are interpreted similarly to a confidence interval in that these intervals quantify the lack of confidence in ML predictions. Unlike confidence intervals, uncertainty intervals only have coverage guarantees if a UQ method is perfectly calibrated, **Fig. 2c**.

*k-fold Ensembling:* The *k*-fold ensemble method estimates uncertainty by segregating the training dataset into *k* nonoverlapping subsets (i.e., folds) that are used to train *k* models. Each model makes adsorption energy predictions on the same test set, but the predictions for these same systems will be different from model to model because each model was trained on a different fold of training data. For every adsorbate/alloy system, $k$ predictions were averaged together to produce a mean adsorption energy prediction $\mu$. The uncertainty (i.e., a standard deviation $\sigma$) was calculated for each average prediction $\mu$. Herein, we used $k = 5$.

*Monte Carlo Dropout:* MC dropout estimates uncertainty by modifying the fully connected layers of the NN model. Dropout was applied both during and after training when making predictions—this protocol is necessary to approximate a Gaussian process [13]. Like the ensemble method, all the adsorption energy predictions corresponding to any specific system were used to calculate a mean adsorption energy prediction $\mu$ and an associated uncertainty $\sigma$ for that system.

We refer to the different adsorption energy predictions for the same adsorbate/alloy system as MC dropout samples. When we say an uncertainty estimate is a 1,000-sample MC dropout estimate, we mean that 1,000 different adsorption energy predictions for the same system were used post-training to calculate a mean prediction $\mu$ and uncertainty estimate $\sigma$. For each fully connected layer of our CGCNN model, 5% of the nodes were stochastically dropped out for each MC dropout sample. A dropout rate of 5% was chosen because it results in a negligible reduction in model accuracy on the test set, **Fig. S1**.

*Evidential Regression:* Evidential regression estimates uncertainty by modifying both the loss function and model architecture of a regression model [50,51]. Evidential regression does not require sampling predictions post-training to estimate the uncertainty, unlike $k$-fold ensembling and MC dropout. All the information needed to make an uncertainty estimate of a prediction is outputted by the model at prediction time. Evidential regression uses an evidential loss function $L_i(w)$ such that the model learns the predictive uncertainty while it minimizes the predictive error during training.

$$L_i(w) = L_i^{NLL}(w) + \lambda L_i^R(w) \quad (1)$$

Here $w$ refers to the model weights that are optimized during training, $L_i^{NLL}(w)$ refers to the negative log-likelihood (*NLL*) of the evidential loss function, and $L_i^R(w)$ refers to the regularization term of the evidential loss function. These terms are defined as follows:

$$L_i^{NLL}(w) = \frac{1}{2}\log\left(\frac{\pi}{v}\right) - \alpha \log(\Omega) \quad (2)$$
$$+ \left(\alpha + \frac{1}{2}\right)\log((y_i - \gamma)^2 v + \Omega)$$
$$+ \log\left(\frac{\Gamma(\alpha)}{\Gamma\left(\alpha + \frac{1}{2}\right)}\right)$$

$$L_i^R(w) = |y_i - \gamma| \cdot (2v + \alpha) \quad (3)$$

Here $\Omega = 2\beta(1 + v)$ and the parameters $\gamma$, $v$, $\alpha$, and $\beta$ are the evidential distribution parameters. The symbol $\Gamma$ refers to the gamma function. The variable $\lambda$ is a scalar hyperparameter that controls the degree of regularization introduced by the term $L_i^R(w)$. We modified the CGCNN model architecture to predict these four parameters for each adsorbate/alloy system. The parameter $\gamma$ represents the predicted target property (here the adsorption energy of a system). The parameters $v$, $\alpha$, and $\beta$ are related to the aleatoric ($\sigma_a$) and epistemic ($\sigma_e$) uncertainties of each prediction:

$$\sigma_a = \frac{\beta}{\alpha - 1} \quad (4)$$

$$\sigma_e = \frac{\beta}{v(\alpha - 1)} \quad (5)$$

Whereas $k$-fold ensembling and MC dropout provide one uncertainty estimate for each prediction, evidential regression provides two estimates corresponding to either aleatoric uncertainty or epistemic uncertainty. For this experiment, we focus on the epistemic uncertainty of evidential regression because epistemic uncertainty is related to the uncertainty of the ML model and is a reducible uncertainty, whereas aleatoric uncertainty is a type of stochastic, irreducible uncertainty [58]. Both aleatoric and epistemic uncertainty estimates across values of $\lambda$ are included in **Fig. S2 and Fig. S3** of the SI for completeness.

The regularization term $L_i^R(w)$ inflates the uncertainty estimate of the adsorption energy predictions based on the absolute residual error $|y_i - \gamma|$, and $\lambda$ weights the magnitude of this inflation. Because $\lambda$ was fixed before training the model and inflates the uncertainty of the predictions, evidential regression gives a tunable uncertainty estimate that grows or shrinks depending on the choice of $\lambda$.

### 2.4 Uncertainty Quantification Metrics

We compare the UQ methods in terms of five UQ metrics: accuracy, sharpness, dispersion, calibration, and tightness [14,43]. These UQ metrics collectively provide a means of comparing the predictive accuracy, shape, and reliability of the uncertainty distribution between UQ methods. Ideally, a UQ method would be accurate, sharp, disperse, calibrated, and tight, **Fig. 2b**. A UQ method that expresses a prediction as a mean and the uncertainty as a standard deviation should give accurate predictions regardless of the uncertainty centered around the predictions. A UQ method is ideally sharp, implying that the method gives highly certain adsorption energy predictions on average. A UQ method is preferably disperse such that the uncertainty level of different test predictions is easily distinguished. Importantly, the UQ method should also be calibrated to allow reliable interpretation as a confidence interval. Lastly, a UQ method should be tight in that the predictive uncertainty interval should only be as large as necessary to capture the ground truth.

*Accuracy:* The mean absolute error (*MAE*), root mean squared error (*RMSE*), median absolute error (*MDAE*), mean absolute relative percent distance (*MARPD*), coefficient of determination ($R^2$), and the correlation coefficient (*R*) were used to quantitatively assess the accuracy of each UQ method. Because *MDAE* uses the median of the absolute residual error distribution across predictions, it is less sensitive to outliers compared to the *RMSE*. *MAE* and *MARPD* are formulated as means, so these metrics have some sensitivity to outliers. *MARPD* is a normalized measure of accuracy [14]:

$$\text{MARPD} = \frac{1}{N}\sum_{i=1}^{N} 100 * \frac{|\hat{y}_i - y_i|}{|\hat{y}_i| + |y_i|} \quad (6)$$

where $\hat{y}_i$ is the $i^{th}$ adsorption energy prediction from the UQ method applied onto our CGCNN model and $y_i$ is the true adsorption energy of the $i^{th}$ adsorbate/alloy system as predicted by DFT.

*Sharpness:* Sharpness (*Sha*) was expressed as [14]:

$$\text{Sha} = \sqrt{\frac{1}{N}\sum_{i=1}^{N} \sigma_i^2} \quad (7)$$

where *N* is the total number of data points and $\sigma_i^2$ is the squared uncertainty of the $i^{th}$ adsorption energy prediction.

*Dispersion:* The dispersion of uncertainty across predictions for each UQ method was measured by constructing a box plot for each method and measuring the interquartile range (*IQR*).

$$\text{IQR} = Q3 - Q1 \quad (8)$$

*Q3* is the third quartile corresponding up to the 75[th] percentile and *Q1* is the first quantile corresponding up to the 25[th] percentile. Whiskers are calculated based on 1.5×*IQR* below the first quartile and above the third quartile, respectively. Each box plot was overlaid with a violin plot, which provides a smooth kernel density estimate of the distribution shape such that each distribution can be visually inspected similarly to a histogram. Each kernel density estimate was performed using Scott's rule for calculating the estimator bandwidth [59].

Coefficient of variation (*Cv*) was calculated to measure dispersion as well. While *IQR* quantifies the spread of a distribution bulk on an absolute scale, *Cv* quantifies the spread of a distribution relative to its mean. These metrics in tandem allow for a more robust analysis of dispersion than either alone. Coefficient of variation was expressed as:

$$\text{Cv} = \frac{\sigma_\sigma}{\mu_\sigma} \quad (9)$$

where $\sigma_\sigma$ and $\mu_\sigma$ respectively are the standard deviation and mean of the uncertainty distribution. We include Bessel's correction in the calculation of $\sigma_\sigma$.

*Calibration:* Calibration formally refers to how well the uncertainty estimate represents the true correctness likelihood of a prediction [38,40–43]. A calibrated ML model is a model whose uncertainty estimates should be comparable with its predictive error residuals [14]. In other words, a calibrated model should often be highly certain for highly accurate predictions and vice versa. Quantitatively, a calibrated model means that the uncertainty magnitude $\sigma$ of an adsorption energy prediction $\mu$ should often be comparable in magnitude to the residual predictive error $(y - \mu)$, where $y$ is the ground-truth adsorption energy of the system the model tries to correctly predict.

Calibration of a model was assessed by constructing a calibration curve, **Fig. 2b**. A calibration curve displays the true frequency of points in each confidence interval relative to the predicted fraction of points in that interval [60]. For calibration curve construction, a quantile-based method was used [55]. Programmatically, this task was accomplished in the same way as explained by Tran *et al* [14].

The residual predictive error was normalized by the uncertainty of that prediction. All the normalized residual errors of all the predictions were combined to construct a distribution. If this distribution was comparable to a unit Gaussian distribution, then the uncertainty across all the predictions was said to be calibrated on average. In other words, the uncertainty across all predictions is calibrated on average if the normalized residual predictive error follows the probability density function of a unit Gaussian distribution:

$$\Phi\left(z = \frac{y - \mu}{\sigma}\right) = \frac{e^{-\frac{z^2}{2}}}{\sqrt{2\pi}} \qquad (10)$$

where $z$ is the normalized residual predictive error. This distribution-specific measure of calibration was applied to all UQ methods tested.

To better interpret this procedure with intuition, consider the case where an adsorption energy prediction is highly accurate but has high uncertainty. In this case, the normalized residual error ($z$) is close to zero because the numerator is small; likewise, the denominator (i.e., uncertainty) is sufficiently large. It is desirable to have a normalized residual less than or equal to unity for predictions because this implicitly shows that the uncertainty interval is large enough to enclose the ground truth adsorption energy. The deviation of the normalized residual error from unit Gaussian behavior is what produces a nonideal calibration curve, which is constructed by defining quantiles and comparing the proportion of data inside each quantile between distributions.

Every calibration curve shown herein visually represents the average calibration, which refers to calibration across the entire adsorption energy test set. However, randomly selected subgroups of predictions and even individual predictions should be calibrated as well [55,60]. We refer to the global scope of calibration across all possible subgroups as the calibration density; to the best of our knowledge, we have not seen this term used in the literature. We use this term to better intuitively emphasize that any stochastic subselection of adsorption energy predictions across varying scales should be calibrated, that is, any subselection of predictions should ideally produce a normalized residual error distribution that is unit Gaussian, **Fig. S4**. If this expectation is met, then we say that a UQ method has effective calibration density. Calibration density is a related concept to individual uncertainty estimates in that it conceptually represents the map of calibration at all sample sizes and quantifies the deviation in the average uncertainty from individual estimates [61].

Calibration density was assessed via adversarial group calibration [55,61]. Ten subgroups of predictions were drawn across different subgroup sizes (e.g., ten subgroups that each represent 10% of the test set, 20%, 30%, etc.). Within each subgroup, predictions were drawn without replacement. For each subgroup size, the ten subgroups were compared in an adversarial fashion. By adversarial, we mean that the worst performing subgroup of the ten compared was chosen as a worst-performing estimate of the calibration density. For each subgroup size, a hundred trials of picking ten subgroups at that size was performed. Using these trials, a mean worst-performing estimate of the calibration density was calculated for each subgroup size, as well as standard error bars (**Fig. 8**).

*Tightness:* It is desirable to score the quality of the uncertainty intervals irrespective of calibration. For example, assessing calibration might indicate that an small uncertainty interval could be sufficient to capture the in-domain ground truth across most predictions, but a UQ method might often be returning unnecessarily larger intervals in practice. The uncertainty intervals should only be as large as necessary to capture the ground truth, so scoring metrics assess what we herein refer to as the tightness of a UQ method. Tightness was measured by computing a negatively oriented mean interval score [39]. An increasingly lower score refers to an increasingly tight and appropriate uncertainty estimate, **Fig. 2c**.

This mean interval score rewards itself for smaller uncertainty intervals that capture the ground truth but punishes itself for smaller uncertainty intervals that do not. Across all predictions, we drew uncertainty intervals across a ground-truth coverage of 1% to 99% in 1% increments. The interval length corresponds to the coverage of our Gaussian distribution assumption (e.g., $\mu \pm 1\sigma$ is a coverage of ~68%, $\mu \pm 2\sigma$ is a coverage of ~95%, etc.). For each prediction, we computed a mean interval score across all coverage levels, **Fig. 2b**. A mean interval score for the entire test set was computed by calculating a mean of these mean scores.

## 2.4 Scalar Recalibration

The UQ methods tested were recalibrated [61] such that the uncertainty interval of $\mu \pm 3\sigma$ more reliably behaves as a confidence interval and encloses the ground truth, which refers to the target property. An interval of ± 3σ was chosen because such an interval provides 99.7% confidence under a perfect calibration assumption. Scalar recalibration was used, which involves multiplying all the uncertainty estimates σ across all adsorption energy predictions by a constant [56]. The constant was chosen via a black-box optimization algorithm that attempts to minimize the miscalibration area for each UQ method, **Table S4**. Brent's method was used for this recalibration method because—unlike Platt scaling [62] for example—it is a nonparametric method that makes no assumptions about the shape of the calibration curve(s) to be recalibrated [63]. The scikit-learn implementation of Brent's method was used [64].

## 3. RESULTS & DISCUSSION

We assess the accuracy of each UQ method applied onto our CGCNN model by comparing parity plots and different quantitative metrics of accuracy, **Fig. 3**. The three UQ methods have similar distributions of adsorption energy predictions. By all quantitative measures, evidential regression outperforms the other UQ methods in accuracy. Evidential regression has an $R^2$ of 0.902 (**Fig. 3c**), whereas MC dropout has 0.893 (**Fig. 3b**) and 5-fold ensemble has 0.890 (**Fig. 3a**). Evidential regression has the lowest *MDAE* of 0.391 eV, suggesting that this UQ method is the most accurate on non-outlier data. Generally, 5-fold ensemble and MC dropout perform similarly across all quantitative metrics of accuracy.

Although each UQ method (**Fig. 3a−c**) differs in its approach for estimating the uncertainty— ensembling involves segregating the training data, MC dropout involves modifying the model architecture, and evidential regression requires modifying both the model architecture and optimization—the accuracy is quite similar for all three methods. To explore why, we examine if the models are overfit or underfit. Consider MC dropout; this method is used to prevent model overfitting as specified by the dropout rate hyperparameter [65]. A higher dropout rate results in dropping out more nodes, thereby resulting in larger adjustments away from overfitting. However, we do not observe improvements in test accuracy as dropout rate increases, suggesting that the baseline CGCNN model that we apply UQ methods onto is not overfitting to the training data—but rather underfitting relative to the chemical space represented by the training data, **Fig. S1**. A plausible explanation for the underfitting of the three UQ methods with a CGCNN is that the OC20 dataset is very sparse; roughly 0.07% of possible calculations were performed when considering the dataset constraints on adsorbates, surfaces, and bulk compositions [33]. Consequently, our CGCNN model architecture may underfit the chemical space represented by the full IS2RE training dataset. To achieve higher accuracy, more state-of-the-art deep neural network models and representations are needed, which is an area of ongoing research [12].

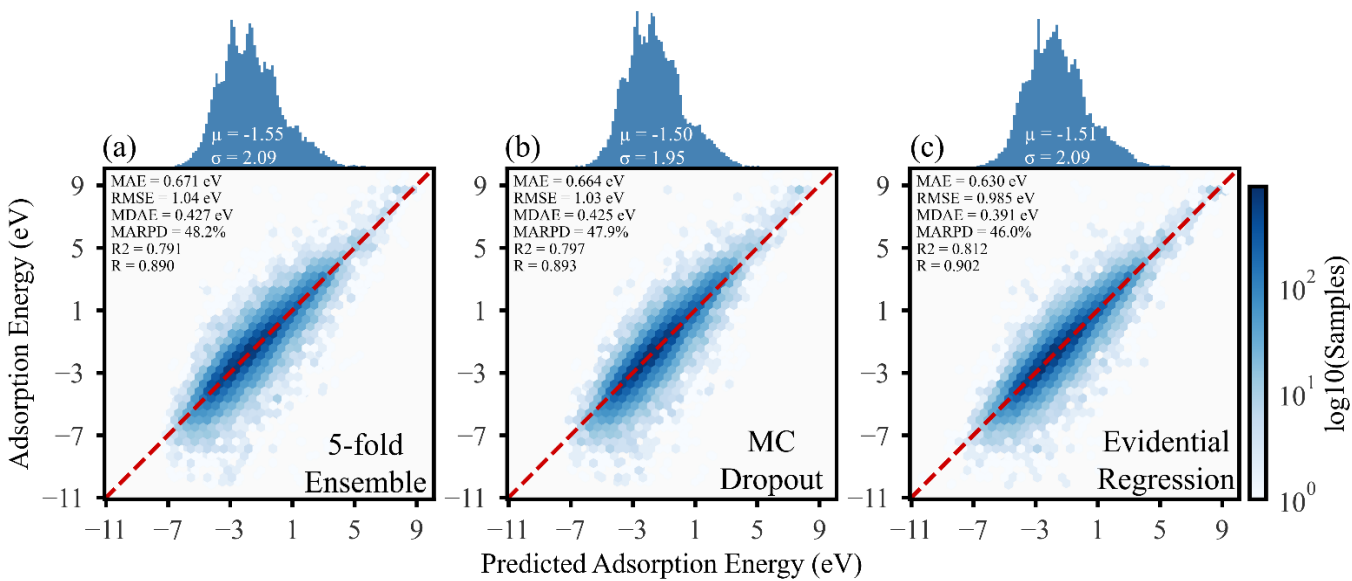

**Fig. 3. Parity plots used to assess accuracy of adsorption energy predictions using CGCNN.** Accuracy results shown for **(a)** 5-fold ensemble, **(b)** 1,000-sample MC dropout, and **(c)** evidential regression with regularization weight λ = 0.05. Hexagonal binning was used. Dashed line (red) is the parity line. The mean absolute error (MAE), root mean squared error (RMSE), median absolute error (MDAE), mean absolute relative percent distance (MARPD), coefficient of determination ($R^2$) and correlation coefficient (R) are reported inset. Histograms of the model predictions are shown on the same scale inset (100 bins) with the mean and standard deviation of the histogram given inset.

We explored the sensitivity of the hyperparameter choice $\lambda$ for evidential regression. **Table S2** contains the test MAE results for evidential regression across varying values of hyperparameter $\lambda$. The test MAE across $\lambda$ fluctuated at most by 0.48% from any given measurement, demonstrating that $\lambda$ can be used to tune the size of the uncertainty intervals for this dataset and model without appreciable changes in the average accuracy of adsorption energy predictions. Based on this sensitivity analysis, we chose $\lambda = 0.05$ for all subsequent experiments because this regularization weight gives the most conservative uncertainty intervals of the non-zero weights tested, **Fig. S3b**.

We construct violin box plots to assess both the dispersion and sharpness of each UQ method, as displayed in **Fig. 4**. High dispersion is desirable because one can define heuristics or rules based on the uncertainty to select and study catalysts for further study. For example, one might want to segregate very uncertain adsorbate/alloy systems from the rest for further study, but there may not be very uncertain systems to segregate if the dispersion is small. MC dropout is the most disperse UQ method on an absolute scale based on having the largest $IQR$. 5-fold ensembling is the least disperse based on having the smallest $IQR$; for this UQ method, defining uncertainty heuristics to select adsorbate/alloy systems for further study could be difficult—at least if one wants to select systems from the bulk of the uncertainty distribution. Evidential regression has dispersion in-between MC dropout and 5-fold ensemble in terms of $IQR$, although we found the dispersion of evidential regression to vary substantially with varying values of $\lambda$, **Fig. S3b**.

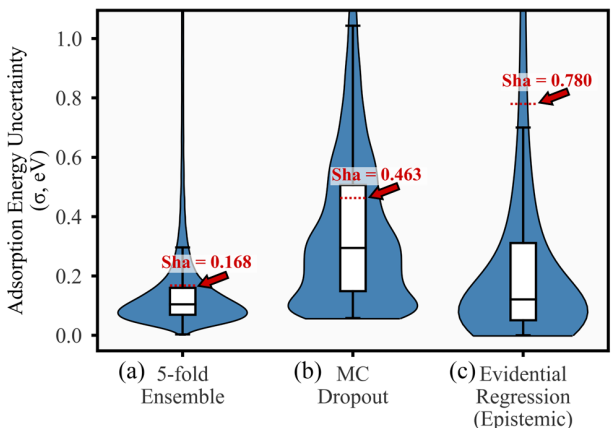

**Fig. 4. Violin box plots to assess dispersion and sharpness for each uncertainty distribution.** Accuracy results shown for **(a)** 5-fold ensemble, **(b)** 1,000-sample MC dropout, and **(c)** evidential regression with $\lambda = 0.05$. Shown inside each box plot: $Q1$ (25th percentile), $Q2$ (50th percentile), $Q3$ (75th percentile), and sharpness (dotted red line). Dispersion as measured by $IQR$ and $Cv$ is reported in **Table S1**. Outliers are not shown for visual clarity of the bulk distribution but are given in **Fig. S5**.

The coefficient of variation analysis demonstrates that the evidential regression results are the most disperse relative to the distribution mean. Evidential regression has a $Cv$ of 2.13, whereas 5-fold ensemble and MC dropout have a $Cv$ of 0.786 and 0.779, respectively. Although MC dropout is the most disperse in absolute terms of the distribution spread, the dispersion of values around the distribution mean is poor. We conclude that evidential regression is the most disperse UQ method based on $IQR$ and $Cv$ collectively.

Of the methods considered, evidential regression is the least sharp (i.e., has the highest sharpness value); thus, for this dataset evidential regression gives the least confident adsorption energy predictions on average. Uncertainty estimates of the adsorption energy predictions can range from 0 eV upwards for each UQ method, so we expected MC dropout to be the least sharp due to having the largest spread as measured by $IQR$, but this is not the case. We observed that evidential regression gives increasingly large—sometimes even enormous—uncertainty estimates for outliers as $\lambda$ increases, which skews the sharpness. Whereas evidential regression is the least sharp (0.780 eV), 5-fold ensemble is the sharpest (0.168 eV). Therefore, the 5-fold ensemble gives the most confident adsorption energy predictions on average. We summarize $IQR$, $Cv$, and sharpness for each UQ method in **Table S1**. Dispersion as measured by $Cv$ for the UQ methods does not change after scalar recalibration. Before recalibration, MC dropout is the most disperse as measured by $IQR$, but this UQ method has poor dispersion as measured by $Cv$. While evidential regression does not have a better $IQR$ before recalibration, this method is generally more disperse before and after recalibration when $IQR$ and $Cv$ are collectively considered. The effects of recalibration are discussed more below.

Although it is important for a UQ method to make confident predictions on average, one needs to ensure that the associated uncertainty estimate $\sigma$ for each adsorption energy prediction are neither overconfident nor underconfident such that we can interpret the uncertainty estimate similarly to that of a confidence interval—an interval that reliably suggests where the actual ground-truth adsorption energy could exist. In other words, it is important to ensure that the UQ method is calibrated.

We assess average calibration across the test set in **Fig. 5** before scalar recalibration. Each calibration curve has an associated miscalibration area, which refers to the area between any given calibration curve and the diagonal. A higher miscalibration area means worse model calibration (i.e., the reliability of using an uncertainty estimate as a confidence interval is weakened). The average calibration between the three UQ methods tested significantly varies, **Fig. 5a**. The data shows that MC dropout provides the most calibrated adsorption energy predictions on average by having the smallest miscalibration area (0.20). Oppositely, 5-fold ensemble gave the least calibrated adsorption energy predictions on average by having the largest miscalibration area (0.38). With a miscalibration area of 0.34, evidential regression performed similarly to the 5-fold ensemble.

We observe a noticeable change in the degree of calibration for MC dropout with an increasing number of samples taken to construct the uncertainty estimate, **Fig. 5b**. 5-sample uncertainty estimates are poorly calibrated, giving a miscalibration area of 0.36. However, 1,000-sample uncertainty estimates reveal that MC dropout is the most calibrated UQ method tested before recalibration. The takeaway is straightforward; one needs to gather enough samples for the uncertainty estimates of MC dropout to converge. For our experimental setup, we found that a sample size of 50 nearly converges MC dropout, **Fig. 5b**. Herein, we report the uncertainty of each adsorption energy prediction using a sample size of 1,000 to ensure convergence.

The data in **Fig. 5c** demonstrates the effect of varying $\lambda$ on the average calibration of evidential regression. Values of 0.0, 0.05, 0.1, 0.15, and 0.2 for $\lambda$ were tested. Evidential regression becomes monotonically more calibrated with increasing $\lambda$ for the values tested. For $\lambda$ values above 0.2, the model struggled to converge onto a finite uncertainty estimate, which we discuss in SI. Interestingly, the calibration curves between all UQ methods were largely of a similar character—most remain below the diagonal. Because of the method of construction, calibration curves below the diagonal indicate overconfident adsorption energy predictions. It is known that increasing $\lambda$ generally inflates the uncertainty estimate of a prediction [50], so it is expected that the evidential regression calibration will improve as overconfident, small uncertainty estimates become appropriately inflated.

If a UQ method does not give calibrated and therefore reliable materials predictions as implemented, the method can be recalibrated. The results in **Fig. 6** show the associated calibration curves on the test set before and after recalibrating each UQ method with scalar recalibration. In general, all three UQ methods saw a sizable decrease in average miscalibration area after recalibration. The data in **Fig. 6a** shows that 5-fold ensembling gives the most calibrated adsorption energy predictions on average based on having the smallest miscalibration area. Despite the highest miscalibration area after recalibration, MC dropout as shown in **Fig. 6b** still gives competitively calibrated adsorption energy predictions on average. The calibration performance of evidential regression displayed in **Fig. 6c** is close to that of MC dropout. All three UQ methods give both overconfident and underconfident uncertainty estimates on average after recalibration, but the degree of miscalibration is less severe.

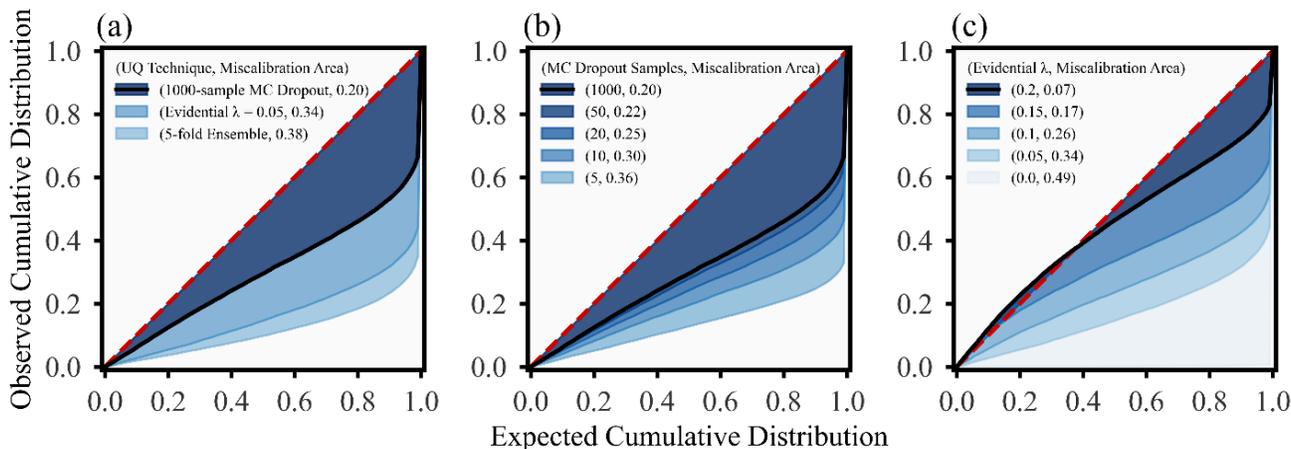

**Fig. 5. Calibration curves to assess average calibration for each UQ method. (a)** Average calibration comparison between 5-fold ensemble, 1,000-sample MC dropout, and evidential regression ($\lambda$ = 0.05). **(b)** The effect of sample size on average calibration of MC dropout. **(c)** The effect of hyperparameter $\lambda$ on average calibration of evidential regression (epistemic uncertainty).

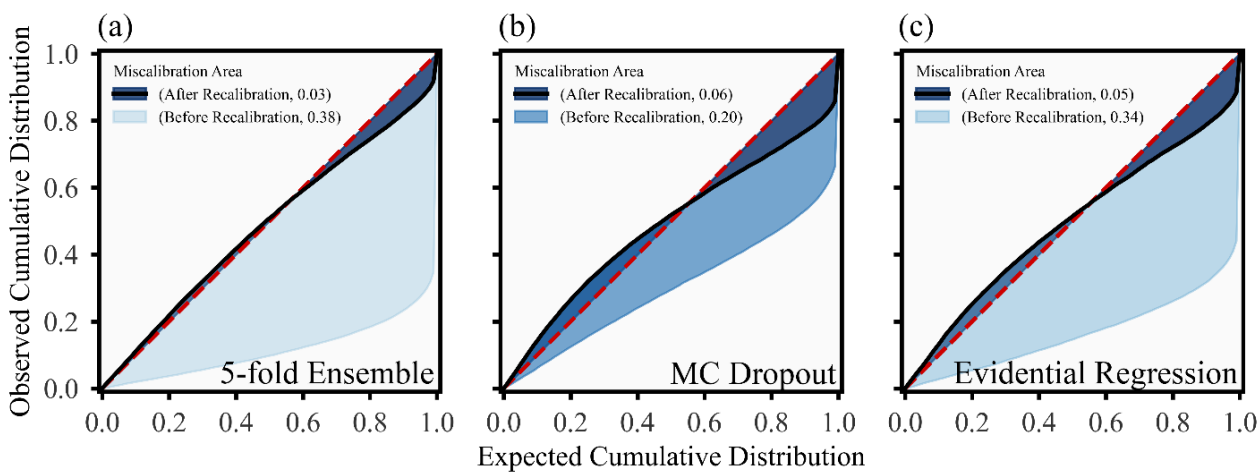

**Fig. 6. Calibration curves to assess recalibration using scalar recalibration.** Calibration curves on the test set are shown for **(a)** 5-fold ensemble, **(b)** 1,000-sample MC dropout, and **(c)** evidential regression ($\lambda$ = 0.05) before and after recalibration. The dashed red line denotes perfect average calibration. The miscalibration area before and after recalibration is shown inset.

All UQ methods we trained have average calibration curves close to the diagonal for the test set after recalibration, which demonstrates that each UQ method is well-calibrated on average for many adsorption energy predictions. Having high calibration on average is useful because selecting adsorbate/alloy systems for further study based on the adsorption energy uncertainty (e.g., in active learning workflows) should be more reliable compared to if the model is highly miscalibrated.

**Table 1** reports the mean interval score to access the tightness of each UQ method before and after recalibration. Before recalibration, evidential regression gives the most appropriately tight uncertainty intervals on average despite not being the most calibrated UQ method. For the predictions that happen to be calibrated, the associated uncertainty intervals of evidential regression are the leanest on average. After recalibration, 5-fold ensemble is the tightest UQ method on average.

**Table 1. Mean interval score for each UQ method.** A lower score indicates better tightness.

| UQ Method | 5-fold Ensemble | 1000-sample MC Dropout | Evidential Regression ($\lambda$ = 0.05) |
|---|---|---|---|
| Interval Score | 5.307 | 4.131 | 3.901 |
| Interval Score (Recalibrated) | 2.907 | 3.472 | 3.550 |

UQ metrics holistically provide a means of comparing the shapes of uncertainty distributions across methods. For example, dispersion quantifies the spread of uncertainty, and sharpness quantifies a notion of average uncertainty. When considering these metrics in their entirety, no UQ method is clearly advantageous before recalibration. MC dropout is the most calibrated on average which might lead one to believe that this method is superior, but evidential regression outperforms the secondary criteria of accuracy, dispersion, and tightness. After recalibration, all UQ methods are comparably calibrated on average, but evidential regression outperforms in measures of accuracy and dispersion. We emphasize that the UQ metrics discussed provide a portfolio from which to holistically compare methods to robustly assess trustworthiness; any single metric on its own can give a misleading conclusion of trustworthiness.

Thus far, we have compared three UQ methods in terms of the five UQ metrics for our challenging benchmark, the OC20 dataset. To demonstrate how these modeling concepts can be used in practice, we report our uncertainty-guided enumeration of materials for a hydrogen adsorbate in **Fig. 7**. We subselect 609 hydrogen adsorbate/catalyst systems from the OC20 Val-ID test dataset used to assess each UQ method and filter these systems based on defined search criteria. Our search criteria for enumerating materials involves defining an adsorption energy range and an uncertainty limit to select systems.

For our case study, we select systems with hydrogen adsorption energies in the range of −0.1 to 0.1 eV. Such a range may be useful, for example, for studying materials for the hydrogen evolution reaction because hydrogen adsorption energy is a descriptor of catalytic activity [66,67]. From these systems, we only choose those with an associated uncertainty $\sigma$ of 0.05 eV or smaller. This adsorption energy range was chosen because each UQ method has inaccuracy due to the performance of the underlying CGCNN model. We decided to select a conservative, narrow search range to select only those predictions that should be confident. With our choice of $\sigma$, a $\mu \pm 3\sigma$ interval implies that the ground-truth adsorption energy prediction should often be at most 0.15 eV away from the ML prediction—the reliability of which becomes less consistent with larger miscalibration. Any material satisfying these criteria is enumerated.

For this case study, our goal is not to rely on our ML adsorption energy predictions for the best chemical accuracy; rather, we demonstrate uncertainty estimates that are at least reliable and trustworthy enough to effectively enumerate materials. Additionally, from a modeling standpoint, we intend to demonstrate the increase in trustworthiness for predictions made with our model on the challenging OC20 dataset before and after recalibration.

We observe improved reliability in the uncertainty estimates after recalibration. Before recalibration, none of the UQ methods enumerated materials that matched our search criteria. After recalibration, evidential regression gave 15 catalyst materials matching the search criteria (**Fig. 7**). Materials that have the same bulk composition differ by adsorption site and Miller index. Crystallographic differences between materials such as $K_2$ and $K_8$ are given in **Table S5**.

A UQ method should propose materials that match the search criteria for an application; however, these proposals should be honest. Six of these materials' predictions are honest in that their associated uncertainty interval of $\mu \pm 3\sigma$ successfully captures the ground truth values. The rest of the proposed materials were dishonest. This lackluster overall screening performance is to be expected using this challengingly sparse OC20 dataset. First and foremost, the trained CGCNN model is not highly accurate as measured by MAE, so many of the adsorption energy predictions are far away from the ground truth values before even applying the UQ methods. A more accurate model would place the predictions closer to the ground truth, demonstrating the importance of recent efforts to develop even more accurate deep NN models [28,32,35,36]. Given that the OC20 dataset is challenging, we note that the CGCNN architecture may have better accuracy on less sparse, narrowly selected datasets, as we observe in **Table S3**. For evidential regression ($\lambda$ = 0.05), we report the element-wise sharpness and accuracy as measured by MAE for catalyst materials in **Fig. S7**, finding that periodic groups 3-5 are particularly uncertain and inaccurate. Additionally, we report uncertainty distributions of evidential regression on a per-adsorbate basis in **Fig. S8**, finding that a handful of adsorbates (e.g., $*NO_2NO_2$) are particularly uncertain for this dataset, model, and UQ method. An improvement in model accuracy should plausibly make any given search criteria more forgiving— assuming that the UQ method applied onto the model is calibrated. We deduce this point from the effective calibration assumption that necessitates that the uncertainty $\sigma$ of any prediction is probabilistically likely to grow or shrink with the accuracy of that prediction.

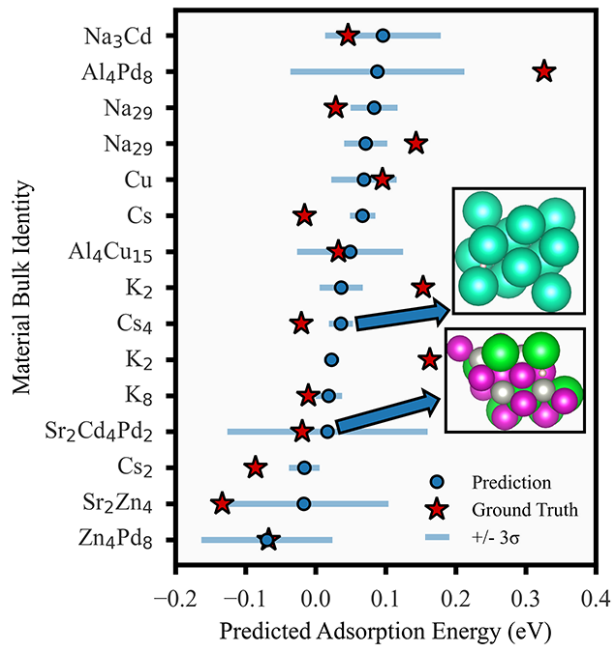

**Fig. 7. Materials compositions from the OC20 dataset that satisfy the enumeration criteria formulated for a hydrogen adsorbate.** Materials were selected using the evidential regression ($\lambda$ = 0.05) uncertainty estimate after scalar recalibration. $Sr_2Cd_4Pd_2$ (Sr: Green, Cd: Purple, Pd: Grey) and $Cs_4$ (Cs: Teal) are visualized for demonstration. Hydrogen adsorbate is not shown. Materials of the same bulk composition differ by adsorption site and Miller index. Details about material differences are given in **Table S5**.

Nevertheless, it is noteworthy that we could identify trustworthy materials predictions as shown in **Fig. 7**, which highlights the importance of recalibration. From a purely data-driven viewpoint, we demonstrate that scalar recalibration did improve a UQ method's ability to enumerate DFT-accurate adsorption energies of these materials, although we acknowledge that many of these materials found in the OC20 dataset are not commonly used catalysts for reactions involving hydrogen. In a real catalyst screening study, one might consider additional criteria outside of adsorption energies, such as stability or selectivity.

Recalibration can be considered as a postprocessing step for a UQ method that can improve an initially poor result, which we demonstrate in this discussion. However, recalibration may not be necessary and may even introduce some drawbacks to the experimental design, depending on the method. For scalar recalibration, all uncertainty estimates $\sigma$ are multiplied by a constant, which has the effect of multiplying the sharpness by said constant. Because sharpness is akin to the average uncertainty of a prediction, scalar recalibration scales the average magnitude of uncertainty, thereby making it more difficult to effectively define narrow $\sigma$ screening criteria if the scaling enlarges the sharpness.

For our demonstration of enumerating hydrogen adsorbate/catalyst systems, we rationalize the increase in the number of enumerated systems after recalibration based on two reasons. First, each UQ method makes overconfident adsorption energy predictions, so inflating all the uncertainty estimates $\sigma$ for any UQ method by a scalar made the model more appropriately calibrated—thereby improving the confidence interval reliability of each uncertainty estimate as supported by **Fig. 7**. Secondly, we define our uncertainty search criteria as $\sigma = 0.05$ eV. Evidential regression ($\lambda = 0.05$) shown in **Fig. 4** is more forgiving for this screening criterion in that the uncertainty estimates $\sigma$ are much less than 0.05 eV before recalibration such that inflating the uncertainty intervals did not exceed the $\sigma = 0.05$ criterion after recalibration, unlike many systems for 5-fold ensemble and MC dropout.

Although the average recalibration results of **Fig. 6** appear excellent and the recalibration did improve the results of our material enumeration case study for the hydrogen adsorbate, a more nuanced analysis of recalibration on the scale of individual predictions was performed by estimating the calibration density. The adversarial group calibration results in **Fig. 8** allow us to estimate the upper bound miscalibration associated with the calibration density across group sizes. We only report here adversarial calibration of group sizes up to 2% of the test set size because we otherwise observe asymptotic miscalibration.

Generally, the miscalibration across UQ methods monotonically improves across group sizes after recalibrating. This consistency in behavior is true near the limiting case of a group size of zero, that is, individual calibration of predictions. However, directly assessing individual calibration is often unverifiable for finite dataset sizes, despite being a strict calibration constraint in the literature [61].

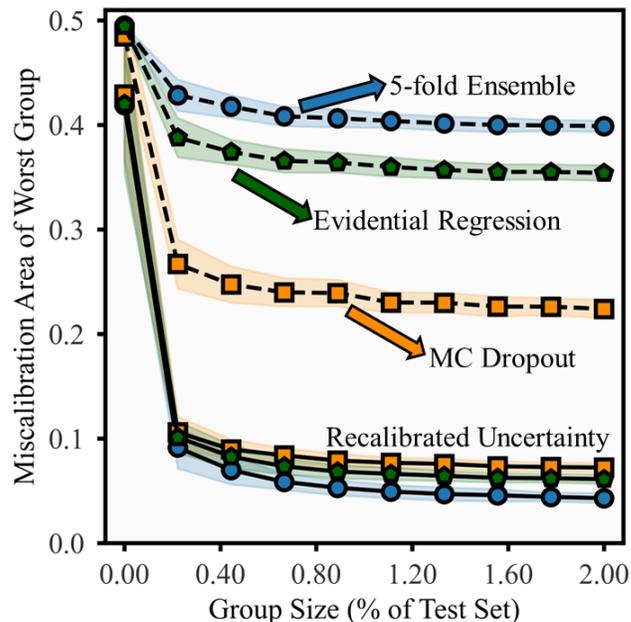

**Fig. 8. Adversarial group calibration to assess calibration density.** The miscalibration area of the most miscalibrated group is shown for that corresponding group size. Shown for 5-fold ensemble (blue circle), 1,000-sample MC dropout (orange square), and evidential regression ($\lambda = 0.05$, green pentagon) before (dotted line) and after (solid line) scalar recalibration. Shaded regions represent the standard error. Trends continue to asymptote past 2.00% group size as shown in **Fig. S6**.

Although we previously calculate average calibration curves across the entire test set, a UQ method is truly calibrated if all individual predictions are themselves calibrated. The recalibration results demonstrate significant improvement in the average calibration (i.e., calibration corresponding to a 100% group size), yet the limiting behavior near the origin in **Fig. 8** suggests that there is only a mild improvement in the individual calibration for the most miscalibrated predictions. Ultimately, the analysis in **Fig. 8** helps rationalize why many of the uncertainty intervals in the **Fig. 7** demonstration do not capture the ground truth. Moreover, we highlight a subtlety that the dramatic average recalibration improvements shown in **Fig. 6** are misleading and, in actuality, more modest, so the interpretation of average calibration needs to be handled carefully for applications.

## 4. CONCLUSIONS

We compare three UQ methods to frame future analyses in high-throughput materials discovery by training a state-of-the-art CGCNN on the challenging OC20 dataset. We base our comparison on the UQ metrics, which quantify not only the accuracy of predictions but also the size, spread, and reliability of the associated uncertainty intervals. Before recalibration, evidential regression is advantageously found to be the most accurate, disperse, and tight UQ method, but MC dropout is the most calibrated method on average. After recalibration, evidential regression is found to be the most accurate and disperse, as well as competitively calibrated. Additionally, evidential regression provides high utility by enabling tunable uncertainty estimates that output upon prediction time, unlike $k$-fold ensembles and MC dropout. This tunability allows for more flexibility in addressing application-specific shortcomings with sharpness, dispersion, and tightness. For future high-throughput studies using distribution-specific UQ, we recommend evidential regression because of its tunability, demonstrated trustworthiness, and computational tractability on this challenging dataset benchmark.

Through the enumeration of OC20 materials in the case of a hydrogen adsorbate, scalar recalibration demonstrably enhances the trustworthiness of evidential regression. We conjecture based on the experimental results that effective recalibration is a promising knowledge gap to make distribution-specific UQ more accessible across materials studies. Neural networks can perform surprisingly poorly under domain shift [68], so it is unclear how each UQ method would perform across different model architectures and complex material spaces. With effective recalibration that has little negative effect on the UQ metrics, the choice of UQ method would be more arbitrary. Researchers would be able to prioritize choosing the method that is the most computationally tractable and recalibrate accordingly. The UQ metrics and analysis discussed can serve as a frame of reference for this endeavor.

## AUTHOR INFORMATION

**Corresponding Author**

*Email: bgoldsm@umich.edu

**Author Contributions**

Conceptualization, C.G., B.R.G.; Methodology, C.G.; Software, C.G.; Formal analysis, C.G., V.M.; Investigation, C.G., V.M.; Resources, C.G., B.R.G.; Data Curation, C.G.; Writing – Original Draft, C.G.; Writing – Review & Editing, C.G., V.M., Y.W., B.R.G.; Visualization, C.G.; Supervision, B.R.G.; Project Administration, B.R.G.; Funding Acquisition, B.R.G.

**Notes**

The authors declare no competing financial interest.

## ACKNOWLEDGMENT

C.G acknowledges support from the NSF Graduate Research Fellowship Program. B.R.G acknowledges support from the MICDE Catalyst Grant from the Michigan Institute for Computational Discovery and Engineering. This research used the National Energy Research Scientific Computing Center resources, a U.S. Department of Energy Office of Science User Facility, operated under Contract No. DE-AC02-05CH11231. This material is based upon work supported by the National Science Foundation Graduate Research Fellowship under Grant No. DGE 1841052.

## ABBREVIATIONS

ML, machine learning; CGCNN, crystal graph convolutional neural network; DER, deep evidential regression; OCP, Open Catalyst Project; OC20, Open Catalyst 20 Dataset; NN, neural network; DFT, density functional theory; MAE, mean absolute error.

# Supporting Information

# Clarifying Trust of Materials Property Predictions using Neural Networks with Distribution-Specific Uncertainty Quantification

Cameron J. Gruich[a,b], Varun Madhavan[a], Yixin Wang[c], and Bryan R. Goldsmith[a,b,*]

[a] Department of Chemical Engineering, University of Michigan, Ann Arbor, Michigan 48109-2136, USA

[b] Catalysis Science and Technology Institute, University of Michigan, Ann Arbor, Michigan 48109-2136, USA

[c] Department of Statistics, University of Michigan, 1085 S University Ave, Ann Arbor, 48109-1107, MI, USA

**Table S1** contains the dispersion and sharpness values across the UQ methods tested before and after scalar recalibration.

**Table S1. Dispersion (measured by $IQR$ and $Cv$) and sharpness (Sha) for each uncertainty quantification method. Values are given in the form of (before recalibration, after recalibration).**

|  | 5-fold Ensemble | 1,000-sample MC Dropout | Evidential Regression ($\lambda = 0.05$) |
|---|---|---|---|
| **Dispersion ($IQR$, eV)** | (0.091, 0.552) | (0.358, 0.832) | (0.260, 1.26) |
| **Dispersion ($Cv$)** | (0.786, 0.786) | (0.779, 0.779) | (2.13, 2.13) |
| **Sharpness (Sha, eV)** | (0.168, 1.02) | (0.463, 1.08) | (0.780, 3.77) |

**Table S2. Evidential regression test mean absolute error results across different values of regularization hyperparameter $\lambda$.**

| $\lambda$ | 0.0 | 0.05 | 0.1 | 0.15 | 0.2 |
|---|---|---|---|---|---|
| **Test MAE (eV)** | 0.633 | 0.630 | 0.632 | 0.632 | 0.633 |

**Table S3** contains values that demonstrate the improvement in test MAE of evidential regression applied to the underlying CGCNN model when the model is trained on solely hydrogen adsorbate systems from the OC20 training dataset, rather than the full training dataset (**Table S2**). We include **Table S3** to highlight that although the CGCNN is not highly accurate on the full OC20 dataset, it can potentially be viable for applications if trained with less sparse chemical spaces than the challenging OC20 dataset.

**Table S3. Evidential regression test accuracy results across different values of regularization hyperparameter $\lambda$ when trained on only systems with a hydrogen (H*) adsorbate from the IS2RE OC20 Training Dataset.**

| $\lambda$ | 0.01 | 0.02 | 0.03 | 0.04 | 0.05 |
|---|---|---|---|---|---|
| Test MAE (eV) | 0.447 | 0.445 | 0.442 | 0.450 | 0.444 |

**Table S4. Recalibration scalars for each UQ method. Scalars were black-box optimized to minimize the average miscalibration area using Brent's method.**

|  | 5-fold Ensemble | 1,000-sample MC Dropout | Evidential Regression ($\lambda$ = 0.05) |
|---|---|---|---|
| Scalar | 6.065 | 2.325 | 4.842 |

**Table S5. Material datapoints that passed the hydrogen adsorbate screening demonstration. Materials with identical bulk identities differ by miller index. Materials with similar bulk identities (e.g., Cs and $Cs_2$) differ by space group.**

| Bulk Identity | OC20 System ID (sid) | Materials Project Bulk ID (mpid) | Miller Index | Crystal System | Symmetry Space Group |
|---|---|---|---|---|---|
| $Na_3Cd$ | sid-431651 | mp-1186031 | (1, 1, 0) | Tetragonal | I4/mmm |
| $Al_4Pd_8$ | sid-564352 | mp-2824 | (0, 1, 0) | Orthorhombic | Pnma |
| $Na_{29}$ | sid-1973951 | mp-1186081 | (2, 1, 0) | Cubic | I$\bar{4}$3m |
| $Na_{29}$ | sid-1811867 | mp-1186081 | (2, 1, 1) | Cubic | I$\bar{4}$3m |
| Cu | sid-2398418 | mp-30 | (1, 1, 0) | Cubic | Fm$\bar{3}$m |
| Cs | sid-1861529 | mp-1183694 | (1, 1, 0) | Tetragonal | I4/mmm |
| $Al_4Cu_{15}$ | sid-1317952 | mp-1228752 | (1, 1, 2) | Tetragonal | P4/mmm |
| $K_2$ | sid-1612615 | mp-972981 | (1, 2, 0) | Orthorhombic | Cmcm |
| $Cs_4$ | sid-695807 | mp-1007976 | (1, 1, 1) | Orthorhombic | Pnma |
| $K_2$ | sid-366931 | mp-972981 | (1, 1, 2) | Orthorhombic | Cmcm |
| $K_8$ | sid-353542 | mp-573691 | (1, 2, 1) | Orthorhombic | Cmce |
| $Sr_2Cd_4Pd_2$ | sid-1534220 | mp-1078825 | (1, 0, 1) | Orthorhombic | Cmcm |
| $Cs_2$ | sid-1799004 | mp-3 | (2, 1, 2) | Tetragonal | I4$_1$/amd |
| $Sr_2Zn_4$ | sid-916136 | mp-569426 | (1, 2, 2) | Orthorhombic | Imma |
| $Zn_4Pd_8$ | sid-411349 | mp-1103252 | (0, 1, 1) | Orthorhombic | Pnma |

The data in **Fig. S1** shows the effect of the dropout rate of fully connected layers on the accuracy of test adsorption energy predictions. For this experimental dataset and hyperparameters, we observed increasing average predictive error as measured by MAE across increasing dropout rates. Beyond a dropout rate of 10%, we observe high growth in MAE, thus we select a dropout rate of 5%.

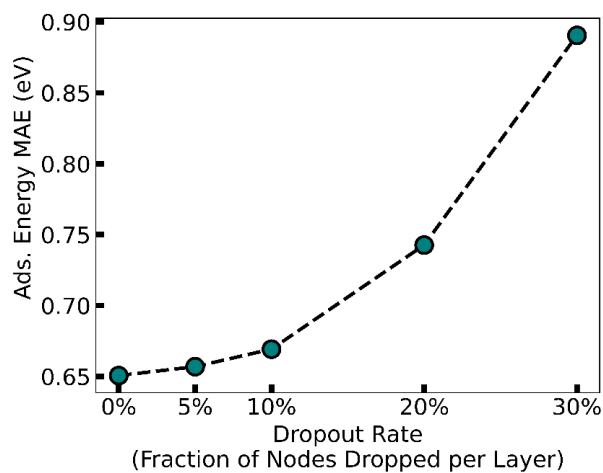

**Fig. S1**. **Effect of dropout rate on the MAE of the predicted adsorption energy on the test dataset changes.**

The data in **Fig. S2** shows the global aleatoric uncertainty trends on the test dataset across different hyperparameter values for evidential regression. The aleatoric uncertainty was highest with no regularization (i.e., $\lambda = 0$). Beyond $\lambda = 0$, the aleatoric uncertainty was regulated to small levels. We do not have strong contentions as to why this happens, but we could reasonably expect low aleatoric uncertainty for this study. The training/test datasets are computationally derived from DFT, which makes them computationally precise and certain from a data viewpoint. Moreover, potential sources of aleatoric uncertainty within the model itself (e.g., stochastic shuffling of data during mini-batch gradient descent) should have relatively little effect on the total uncertainty compared to epistemic sources, such as growing or shrinking the amount of CGCNN weights. For example, if the random shuffling in mini-batch gradient descent dramatically and consistently changed the test accuracy, then neural networks themselves would optimize chaotically for any given fixed model and test set, greatly nullifying the ability of these models to automatically learn patterns in the data.

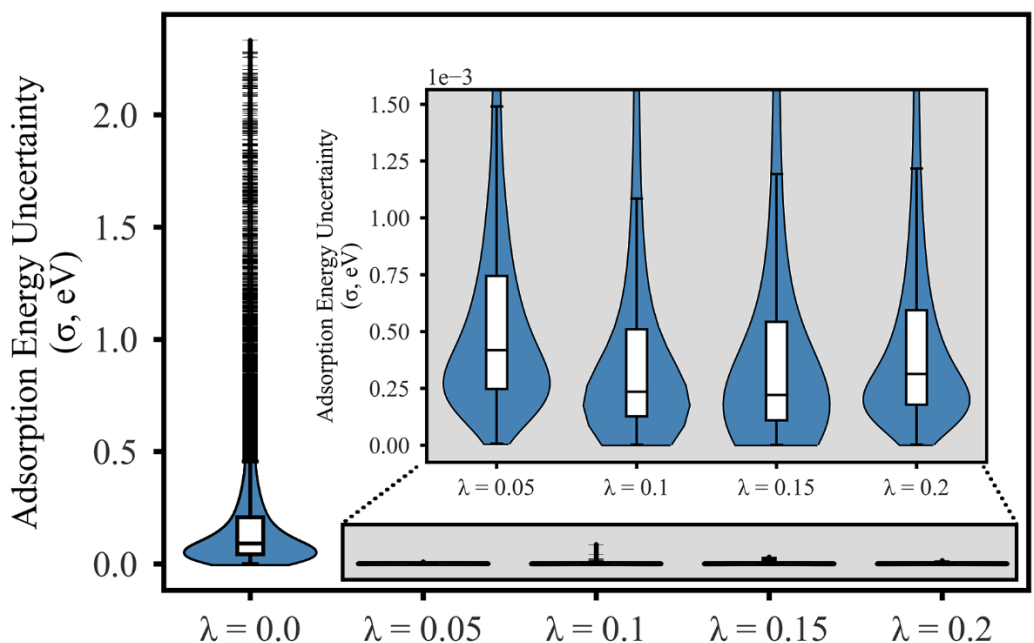

**Fig. S2. Global aleatoric uncertainty trend across different values of hyperparameter $\lambda$ for evidential regression.** Outliers (e.g., points below $Q1 - 1.5 \times IQR$ or above $Q3 + 1.5 \times IQR$) are shown as horizontal black lines.

**Fig. S3a** shows the global epistemic uncertainty trend across different values of $\lambda$ for evidential regression. The uncertainty outliers are generally distanced further away from the bulk of the distribution with increasing values of $\lambda$, but this is not monotonic. We note that values of $\lambda$ tested beyond 0.2 resulted in convergence issues of the model during the gradient descent process. However, we also note that the uncertainty intervals become impractically large as $\lambda$ increases. We only observed convergence issues in these extreme scenarios where $\lambda$ gave impractically large uncertainty intervals. **Fig. S3b** demonstrates that the dispersion of evidential regression as measured by $IQR$ increases with increasing $\lambda$. The distance between the lower whisker and the median grows slowly compared to the distance between the median and the upper whisker across $\lambda$, implying that roughly half the data stays in a relatively low regime of uncertainty even with the increase in $\lambda$.

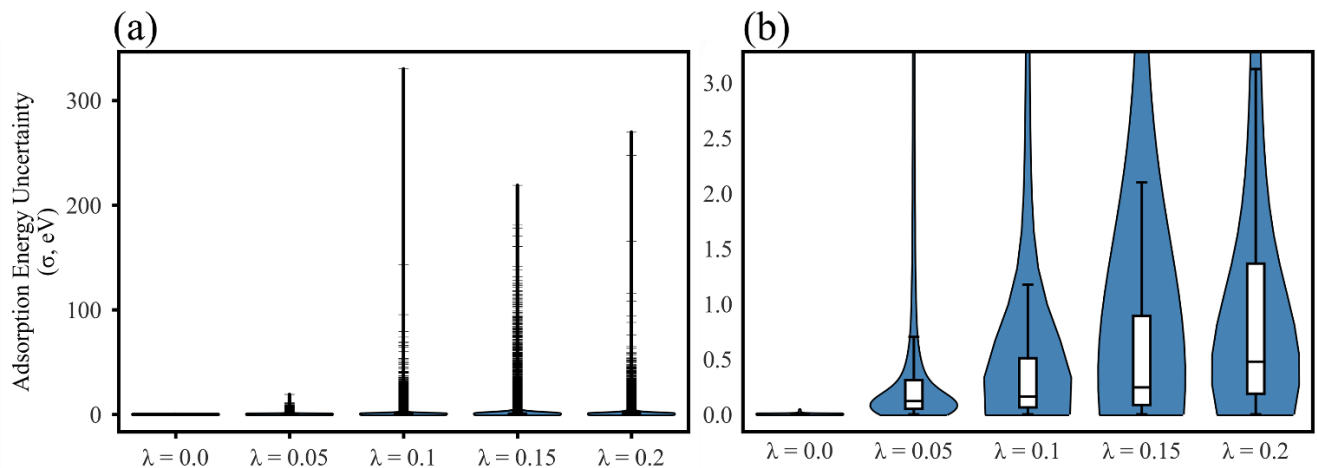

**Fig. S3. Epistemic uncertainty trend for evidential regression. (a)** Global epistemic uncertainty trend across different values of $\lambda$. **(b)** Zoomed-in epistemic uncertainty trend across different values of $\lambda$.

**Fig. S4** illustrates the differences between average, group, and individual calibration, which collectively form a map of calibration density. The average calibration refers to the calibration curve that is constructed using the entire test set of predictions. Group calibration refers to the calibration curve that is constructed using a subset of the test set of predictions. Individual calibration refers to the calibration curve that is constructed using a single test prediction. **Fig. S4** highlights the importance of calibration density in distribution-specific UQ methods in that the average calibration may look promising, but the calibration of individual predictions might be quite bad.

Computing a standard deviation with one data point is problematic, so one strategy to assess individual calibration is to adversarially compare group calibration for smaller and smaller group sizes as the limit is taken to one prediction. Different combinations of groups in **Fig. S4** can be drawn; the combinatoric space of possible groups for a given group size can be massive. To overcome this challenge, groups can be adversarially compared. For example, a million groups of two predictions can be compared to each other, and the worst-calibrated group can be used as a lower-bound estimate for the degree of individual calibration. We refer to the entire combinatoric space of groups as the calibration density, which includes both the entire test dataset and individual predictions as groups. A distribution-specific UQ method that is calibrated on all levels (i.e., average, group, individual) is said to have effective calibration density by our definition. This definition is largely semantic rather than methodological, and it is defined to better communicate this calibration challenge.

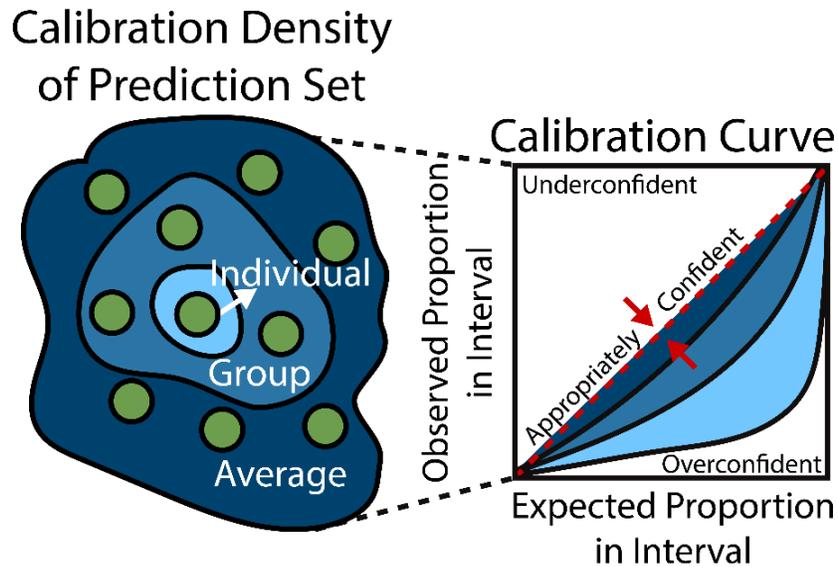

**Fig. S4. Illustration of calibration density.** Average, group, and individual calibration curves are shown on the same plot to demonstrate how calibration can change across data subsets. Shaded regions on the left correspond to shaded regions on the calibration curves.

S2

The data in **Fig. S5** shows the global trend for epistemic uncertainty across the UQ methods tested. The bulk of the epistemic distribution for evidential regression appears thinner than MC dropout, but outliers are comparably more uncertain. **Fig. S5** suggests that evidential regression segregates and inflates uncertain outliers more than MC dropout. This quality can be beneficial or harmful, depending on the viewpoint. Segregation of outliers can be beneficial in that uncertain systems can be identified for active learning improvement. For example, more accurate DFT calculations on these uncertain material outliers could be calculated and recycled back into the training dataset. The model can potentially be retrained in this way to improve chemical space generalization. However, not refining the ML model based on these outliers and accepting the state of the model as-is is problematic because these outliers inflate the sharpness metric that one relies on to quantify trustworthiness.

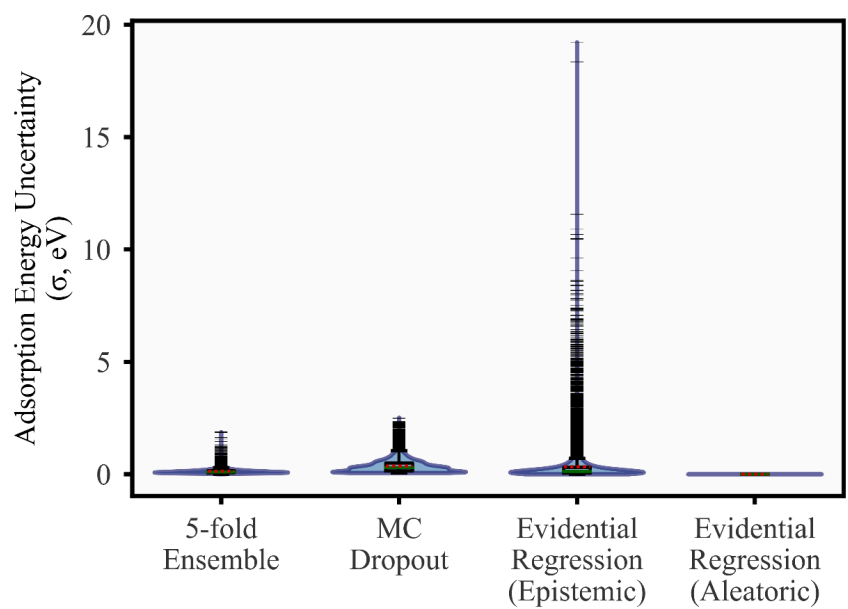

**Fig. S5. Global trends for uncertainty across the UQ methods tested.** Shown for 5-fold ensemble, 1,000-sample MC dropout, and evidential regression ($\lambda = 0.05$). Outliers (e.g., points below $Q1 - 1.5 \times IQR$ or above $Q3 + 1.5 \times IQR$) are shown as horizontal black lines.

The data in **Fig. S6** shows the global trends for adversarial group calibration with group sizes up to 100% before and after recalibration. **Fig. 8** suggests that the trends begin to asymptote beyond a group size of 2%, and **Fig. S6** shows this explicitly. We contend that we reasonably estimated calibration density for this experimental setup because the trends are asymptotic with a low standard error across a wide range of group sizes.

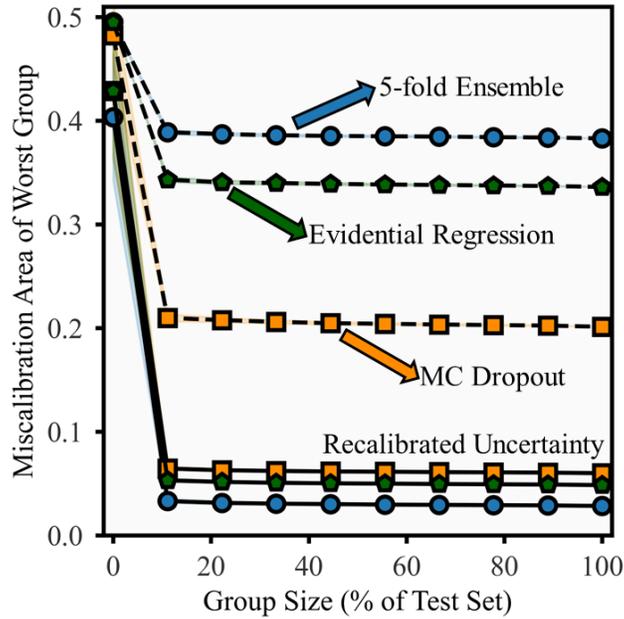

**Fig. S6. Global trend for adversarial group calibration to assess calibration density for the three UQ methods.** The miscalibration area of the worst performing group is shown for that corresponding group size. Shown for 5-fold ensemble (blue circles), 1,000-sample MC dropout (orange squares), and evidential regression ($\lambda$ = 0.05, green pentagons) before (dotted line) and after (solid line) scalar recalibration. Shaded regions represent the standard error.

**Fig. S7** shows the element-wise sharpness and accuracy of catalyst materials corresponding to adsorption energy predictions from the IS2RE Val-ID dataset. We note that periodic groups 3-5 are more uncertain and inaccurate relative to other groups. The uncertainty magnitude seems to correlate decently well with the accuracy upon visual inspection. In practice, one could explore excluding materials containing uncertain elements (e.g., Hf) to potentially improve the model accuracy, which could improve any downstream uncertainty screening in high-throughput experimental design.

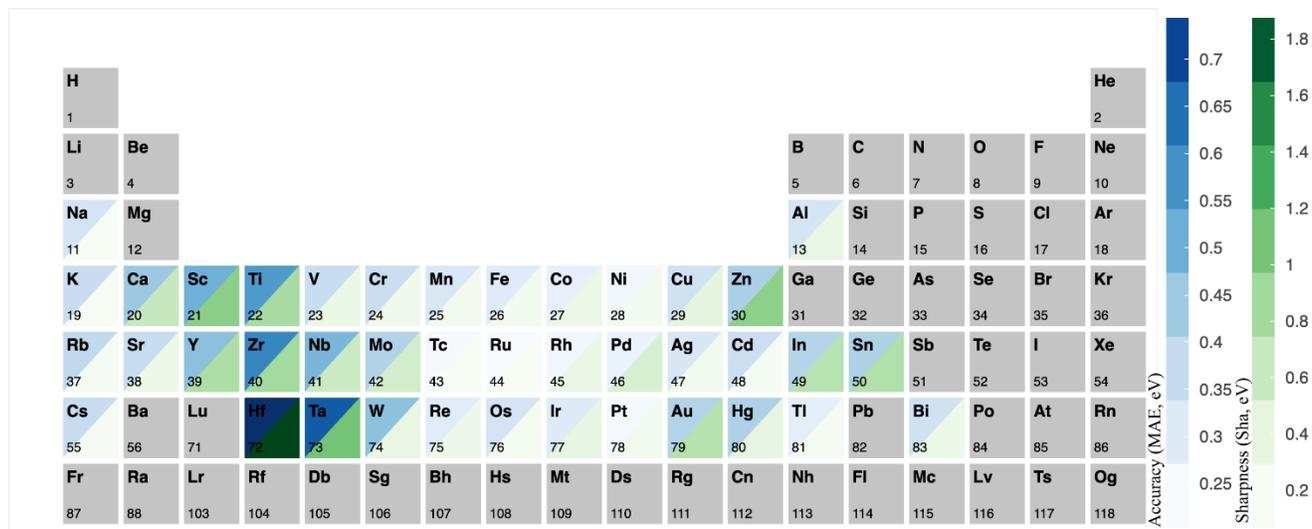

**Fig. S7.** Element-wise sharpness (green) and accuracy (MAE, blue) of catalyst materials for evidential regression ($\lambda = 0.05$) adsorption energy predictions before recalibration from the IS2RE Val-ID dataset.

S5

**Fig. S8** shows the per-adsorbate uncertainty distributions corresponding to adsorption energy predictions from the IS2RE Val-ID dataset and suggests particularly uncertain adsorbates. The adsorbate *NO$_2$NO$_2$ is the most uncertain by having the most disperse box plot. In principle, one could remove *NO$_2$NO$_2$ systems to improve the model accuracy and downstream uncertainty screening for a high-throughput application. Other adsorbates were quite uncertain relative to others as well, such as *NO$_3$, *NO$_2$, and *CCH$_2$.

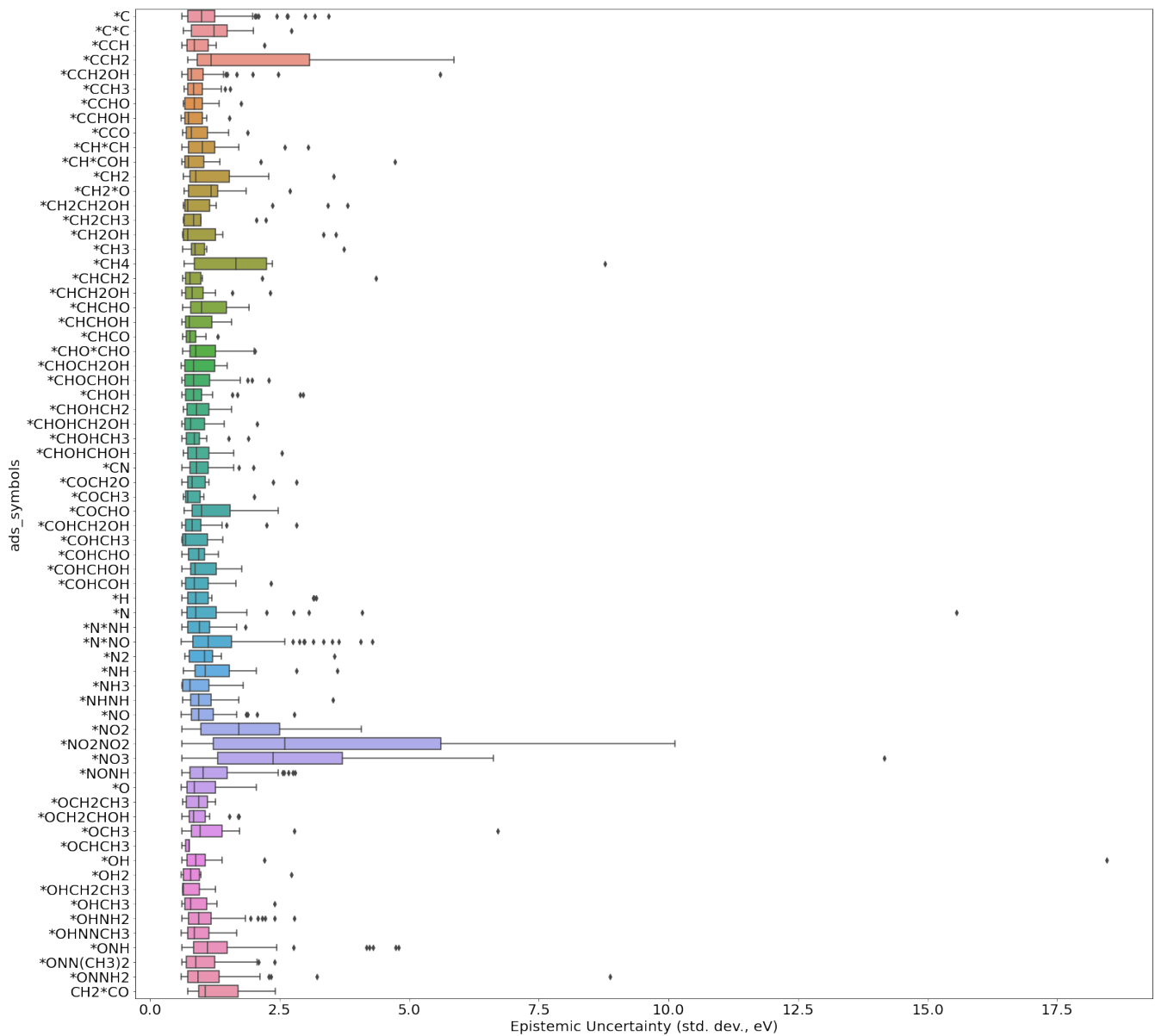

**Fig. S8. Per-adsorbate epistemic uncertainty distributions for evidential regression ($\lambda$ = 0.05) adsorption energy predictions on the IS2RE Val-ID dataset.**